\documentclass{article}

\usepackage{arxiv}

\usepackage[utf8]{inputenc}
\usepackage[T1]{fontenc}
\usepackage{hyperref}
\usepackage{url}
\usepackage{booktabs}
\usepackage{amsfonts}
\usepackage{nicefrac}
\usepackage{microtype}
\usepackage{graphicx}
\usepackage{float}
\usepackage{amsmath,amssymb,amsfonts,amsthm}
\usepackage{multirow}
\usepackage{array}
\usepackage{xcolor}
\usepackage{caption}
\usepackage{gensymb}
\usepackage{enumitem}
\usepackage{subfig}
\usepackage{cleveref}
\usepackage{setspace}
\usepackage{natbib}
\usepackage{doi}
\usepackage{placeins}
\usepackage{adjustbox}

\title{A Novel Hybrid PID-LQR Controller for Sit-to-Stand Assistance
Using a CAD-Integrated Simscape Multibody Lower Limb Exoskeleton}

\author{
  Ranjeet~Kumbhar \\
  Department of Mechanical Engineering\\
  Thapar Institute of Engineering and Technology\\
  Patiala, India \\
  \texttt{ranjeet.kumbhar@thapar.edu} \\
  \And
  Rajmeet~Singh\thanks{Corresponding author}  \\
  Khalifa University Center for Autonomous Robotic Systems (KUCARS)\\
  Khalifa University\\
  Abu Dhabi, UAE \\
  \texttt{rajmeet.bhourji@ku.ac.ae} \\
  \And
  Appaso~M~Gadade \\
  Department of Mechanical Engineering\\
  Thapar Institute of Engineering and Technology\\
  Patiala, India \\
  \texttt{appaso.gadade@thapar.edu} \\
  \And
  Ashish~Singla \\
  Department of Mechanical Engineering\\
  Thapar Institute of Engineering and Technology\\
  Patiala, India \\
  \texttt{ashish.singla@thapar.edu} \\
  \And
  Irfan~Hussain\\
  Khalifa University Center for Autonomous Robotic Systems (KUCARS)\\
  Khalifa University\\
  Abu Dhabi, UAE \\
  \texttt{irfan.hussain@ku.ac.ae} \\
}

\renewcommand{\shorttitle}{Hybrid PID-LQR for STS Exoskeleton Assistance}

\hypersetup{
  pdftitle={A Novel Hybrid PID-LQR Controller for Sit-to-Stand Assistance},
  pdfsubject={Robotics, Control Systems, Rehabilitation Engineering},
  pdfauthor={Ranjeet Kumbhar, Rajmeet Singh, Appaso M Gadade,
             Ashish Singla, Irfan Hussain},
  pdfkeywords={Lower limb exoskeleton, sit-to-stand, Hybrid PID-LQR,
               Simscape Multibody, OpenSim, Rehabilitation robotics,
               Trajectory tracking},
}

\begin{document}
\maketitle

\begin{abstract}
Precise control of lower limb exoskeletons during sit-to-stand (STS)
transitions remains a central challenge in rehabilitation robotics owing
to the highly nonlinear, time-varying dynamics of the human-exoskeleton
system and the stringent trajectory tracking requirements imposed by
clinical safety. This paper presents the systematic design, simulation,
and comparative evaluation of three control strategies, a classical
Proportional-Integral-Derivative (PID) controller, a Linear Quadratic
Regulator (LQR), and a novel Hybrid PID-LQR controller applied to
a bilateral lower limb exoskeleton performing the sit-to-stand
transition. A high-fidelity, physics-based dynamic model of the
exoskeleton is constructed by importing a SolidWorks CAD assembly
directly into the MATLAB/Simulink Simscape Multibody environment,
preserving accurate geometric and inertial properties of all links.
Physiologically representative reference joint trajectories for the
hip, knee, and ankle joints are generated using OpenSim musculoskeletal
simulation and decomposed into three biomechanical phases:
flexion-momentum (0--33\%), momentum-transfer (34--66\%), and extension
(67--100\%). The proposed Hybrid PID-LQR controller combines the optimal
transient response of LQR with the integral disturbance rejection of
PID through a tuned blending coefficient $\alpha = 0.65$. Simulation
results demonstrate that the Hybrid PID-LQR achieves RMSE reductions of
72.3\% and 70.4\% over PID at the hip and knee joints, respectively,
reduces settling time by over 90\% relative to PID across all joints,
and limits overshoot to 2.39\%--6.10\%, confirming its superiority over
both baseline strategies across all evaluated performance metrics and
demonstrating strong translational potential for clinical assistive
exoskeleton deployment.
\end{abstract}


\section{Introduction}

The sit-to-stand (STS) and stand-to-sit (StS) transitions are among the
most biomechanically demanding activities of daily living, requiring the
precise coordination of multiple lower limb joints against gravitational
loading \citep{b1}. For individuals affected by neurological impairments, including spinal cord injury, stroke, and age-related motor decline
these transitions frequently represent the boundary between
independence and functional dependence. Exoskeletons have emerged as a
powerful means of restoring or augmenting the capacity for such
transitions, yet the quality of assistive motion is ultimately determined
by the sophistication of the underlying control strategy.

Among classical approaches, PID control is widely adopted for its
conceptual simplicity and ease of implementation, making it a widely used
solution for exoskeleton joint trajectory tracking \citep{b2,b3}.
However, PID is an inherently local, single-variable approach, tuned
independently at each joint without systematically accounting for the
global energy or state of the coupled human-exoskeleton system
\citep{b4}. The Linear Quadratic Regulator (LQR) addresses this
limitation through optimal state-feedback design, providing a
mathematically principled trade-off between tracking error minimization
and control torque minimization \citep{b5,b6,b7}. Yet each carries
inherent trade-offs: PID requires manual gain tuning and may produce
steady-state error under load; LQR relies on a linearised model and may
lack robustness to modeling uncertainties and the highly variable
inertial parameters arising from user variability \citep{b8}. A Hybrid
PID-LQR architecture offers an attractive synthesis leveraging LQR's
optimal state-feedback structure while incorporating PID's integral action
to eliminate steady-state error and compensate for unmodeled dynamics
\citep{b9}. Despite extensive exoskeleton control literature, a direct
and systematic comparison of PID, LQR, and a hybrid PID-LQR architecture
applied specifically to STS transitions remains absent.

\noindent The principal contributions of this paper are as follows:
\begin{itemize}[leftmargin=*, itemsep=2pt]
  \item \textbf{CAD-integrated Simscape Multibody model:} A high-fidelity,
        physics-based dynamic model of the bilateral lower limb exoskeleton
        is developed by importing a detailed SolidWorks CAD assembly into
        MATLAB/Simulink Simscape Multibody, preserving accurate geometric
        and inertial properties.
  \item \textbf{OpenSim-derived reference trajectories:} Physiologically
        representative STS joint trajectories are generated via OpenSim
        inverse kinematics across three biomechanical phases.
  \item \textbf{Novel Hybrid PID-LQR controller:} A hybrid architecture
        exploiting the complementary strengths of PID and LQR is proposed
        as the principal contribution for STS exoskeleton assistance.
  \item \textbf{Systematic multi-metric comparison:} A rigorous evaluation
        against PID and LQR baselines across tracking accuracy, transient
        response, and steady-state error metrics is conducted.
\end{itemize}

\section{Methodology}

\subsection{Simulation Framework and Exoskeleton Model}

\begin{figure}[H]
  \centering
  \includegraphics[width=0.90\linewidth]{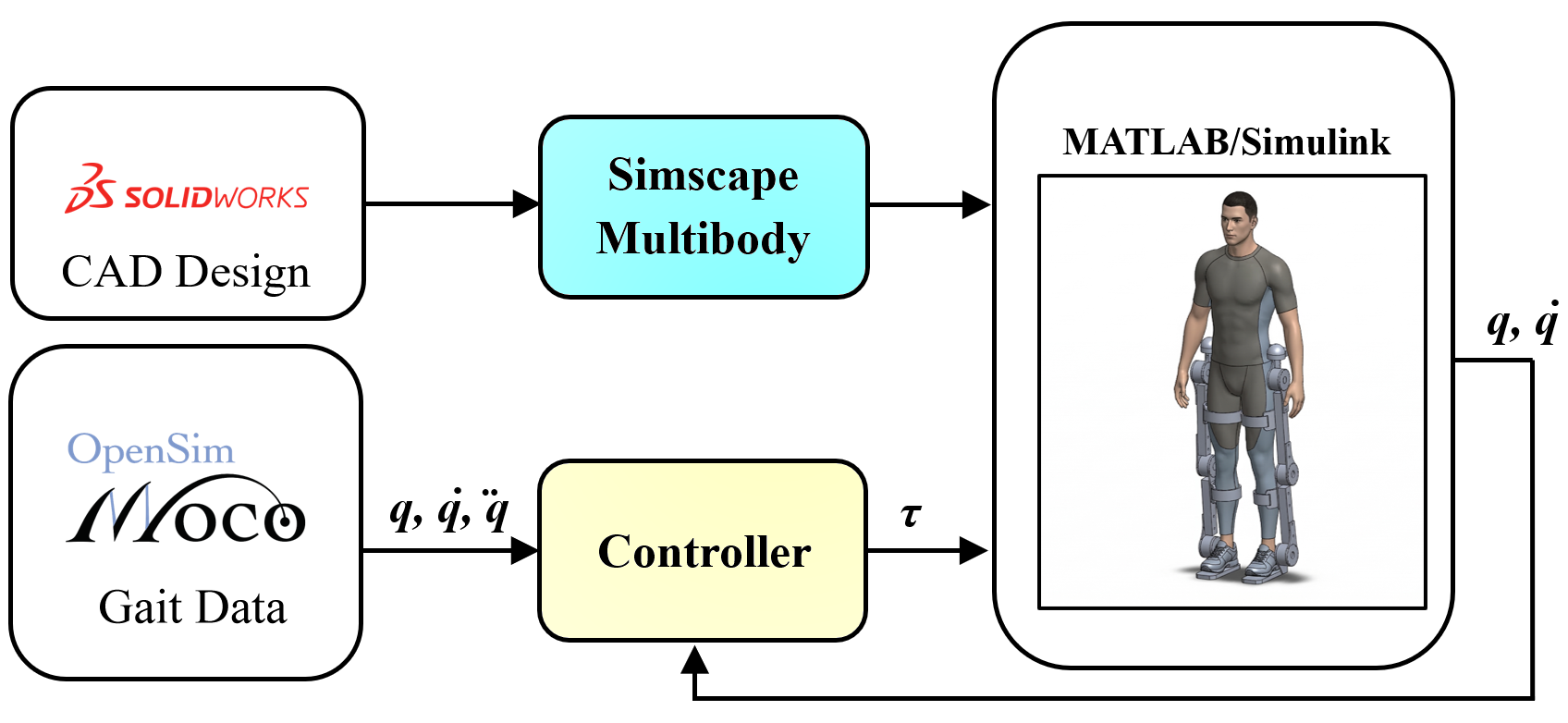}
  \caption{Overall simulation framework for the exoskeleton control
  system. The mechanical model is designed in SolidWorks and imported
  into MATLAB/Simulink via Simscape Multibody, while joint reference
  trajectories ($q$, $\dot{q}$, $\ddot{q}$) are derived from OpenSim
  Moco gait data. The controller receives reference trajectories
  alongside feedback states ($q$, $\dot{q}$) from the Simscape model
  and computes the required joint torques ($\tau$) to drive the
  human-exoskeleton system through the STS transition.}
  \label{fig:fig1}
\end{figure}

\begin{figure}[H]
  \centering
  \includegraphics[width=\linewidth]{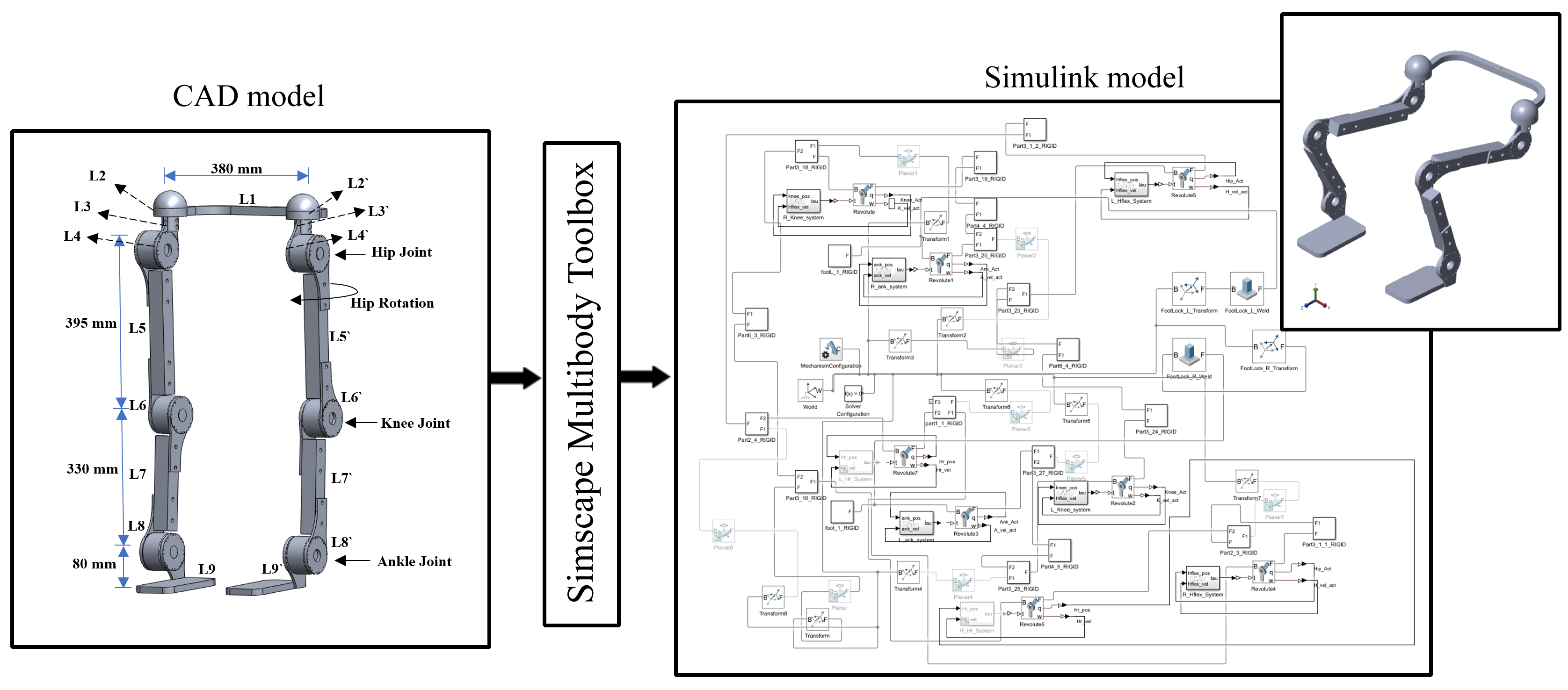}
  \caption{CAD-to-Simulink Simscape Multibody workflow for the bilateral
  lower limb exoskeleton. (\textit{Left}) SolidWorks CAD assembly showing
  the bilateral kinematic chain with labeled rigid links L1--L9 and
  annotated revolute joints at the hip, knee, and ankle. (\textit{Center})
  Simscape Multibody Link toolbox import interface. (\textit{Right})
  Auto-generated Simscape Multibody block diagram with the Mechanics
  Explorer 3-D rendering inset.}
  \label{fig:fig2}
\end{figure}

Figure~\ref{fig:fig1} depicts the overall simulation framework. The
exoskeleton CAD assembly (SolidWorks, Al 6061-T6 material) is imported
into MATLAB/Simulink Simscape Multibody via the Simscape Multibody Link
workflow \citep{b21,b26,b27}, which exports an XML topology file and STL
mesh geometry loaded via the \texttt{smimport()} function. This approach
automatically preserves rigid body masses, inertia tensors, and revolute
joint topology, eliminating the manual derivation overhead of the
Lagrangian formulation for a bilateral six-DOF system:
\begin{equation}
\mathbf{M}(\mathbf{q})\ddot{\mathbf{q}} +
\mathbf{C}(\mathbf{q},\dot{\mathbf{q}})\dot{\mathbf{q}} +
\mathbf{G}(\mathbf{q}) = \boldsymbol{\tau}
\label{eq:EOM}
\end{equation}
where $\mathbf{M}(\mathbf{q}) \in \mathbb{R}^{n \times n}$ is the
configuration-dependent inertia matrix, $\mathbf{C}(\mathbf{q},
\dot{\mathbf{q}}) \in \mathbb{R}^{n \times n}$ is the Coriolis matrix,
$\mathbf{G}(\mathbf{q}) \in \mathbb{R}^{n}$ is the gravity vector,
$\boldsymbol{\tau} \in \mathbb{R}^{n}$ is the joint torque vector, and
$\mathbf{q} \in \mathbb{R}^{n}$ denotes the generalized coordinates with
$n=6$ for the six active joints (hip, knee, ankle per side)
\citep{b13,b15}. The CAD-to-Simscape workflow is illustrated in
Figure~\ref{fig:fig2}, and the physical link parameters extracted
directly from the CAD assembly are listed in Table~\ref{tab:table1}.

\begin{table}[H]
  \centering
  \caption{Physical parameters of the lower limb exoskeleton links
  extracted from the CAD-integrated Simscape Multibody model
  (Al 6061-T6, $\rho = 2700$\,kg/m$^3$).}
  \label{tab:table1}
  \renewcommand{\arraystretch}{1.2}
  \setlength{\tabcolsep}{5pt}
  \begin{tabular}{clccc}
    \toprule
    \textbf{Link} & \textbf{Segment} & \textbf{L\,(mm)} &
    \textbf{m\,(kg)} & \textbf{CoM\,[x,y,z]\,(m)} \\
    \midrule
    L1 & Backboard          & 380 & 0.358 & $[0,\ 0.010,\ 0.086]$      \\
    L2 & Hip rot. bracket   & --- & 0.182 & $[0,\ 0.059,\ 0]$          \\
    L3 & Hip flex connector & --- & 0.054 & $[0.003,\ 0.011,\ 0.003]$  \\
    L4 & Hip joint conn.    & --- & 0.068 & $[0.003,\ 0.030,\ 0.007]$  \\
    L5 & Thigh link         & 395 & 0.459 & $[0,\ 0.148,\ 0]$          \\
    L6 & Knee joint conn.   & --- & 0.068 & $[0.003,\ 0.030,\ 0.007]$  \\
    L7 & Shank link         & 330 & 0.347 & $[0,\ 0.113,\ 0]$          \\
    L8 & Ankle joint conn.  & --- & 0.068 & $[0.003,\ 0.031,\ 0.007]$  \\
    L9 & Foot               &  80 & 0.366 & $[0.046,\ 0.064,\ -0.05]$  \\
    \bottomrule
  \end{tabular}
\end{table}

\subsection{Gait Generation for Sit-to-Stand Operation}

\begin{figure}[H]
  \centering
  \includegraphics[width=\linewidth]{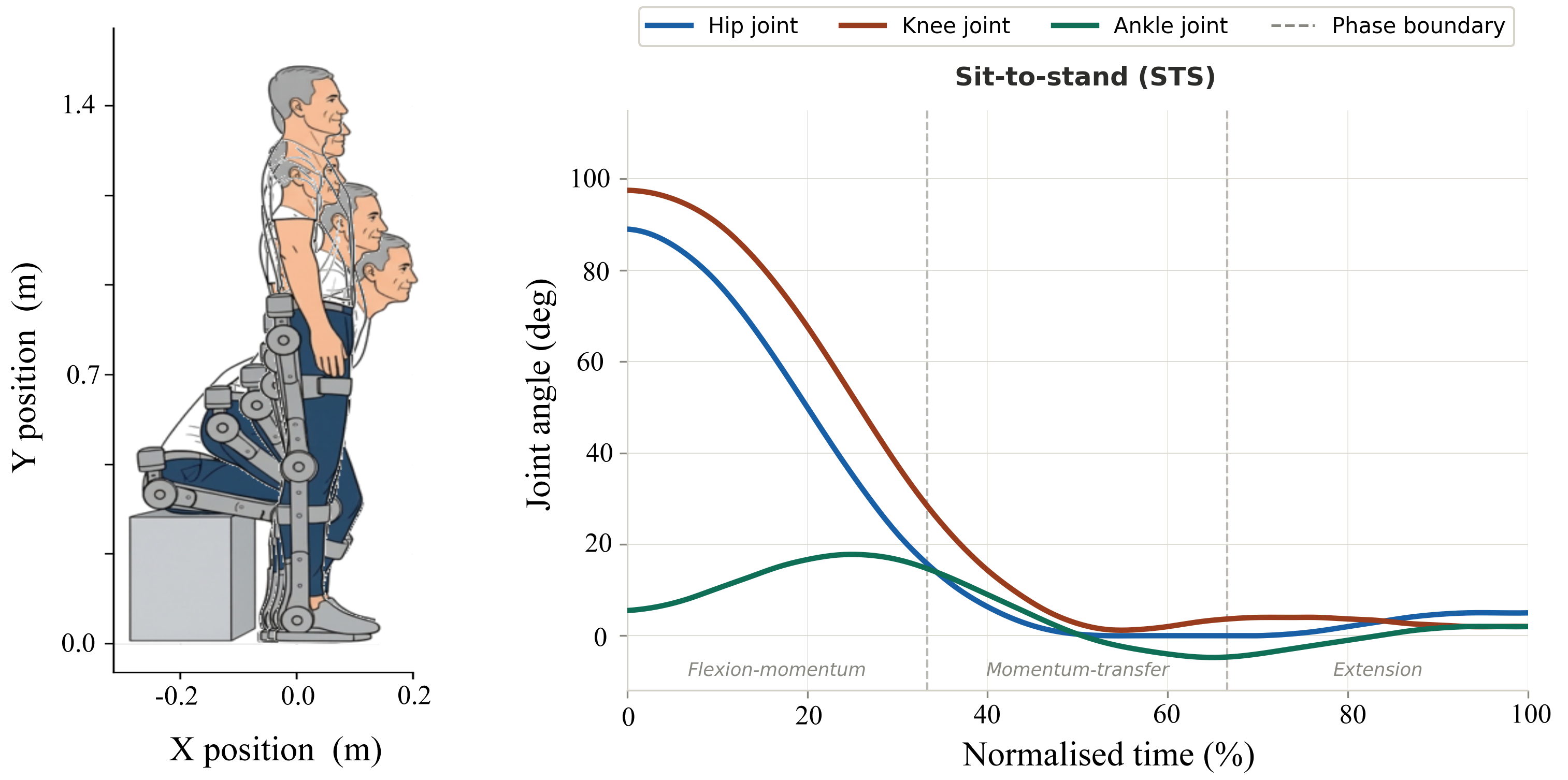}
  \caption{Sit-to-stand (STS) gait generation. (\textit{Left}) Simscape
  Multibody simulation showing sequential postural configurations of the
  human-exoskeleton system from seated to full upright stance in the
  sagittal plane. (\textit{Right}) Reference joint angle trajectories for
  the hip, knee, and ankle joints over normalised motion time (0--100\%,
  total duration 3.0\,s), with dashed vertical lines demarcating the
  three biomechanical phases: flexion-momentum (0--33\%),
  momentum-transfer (34--66\%), and extension (67--100\%).}
  \label{fig:fig3}
\end{figure}

Reference joint trajectories are generated using the OpenSim
musculoskeletal simulation environment \citep{b22} for a 50th-percentile
reference subject (75\,kg, 1.75\,m). The joint angle profiles are
low-pass filtered at 6\,Hz using a fourth-order zero-lag Butterworth
filter, normalised to a uniform motion duration of 3.0\,s, and resampled
at 1\,kHz. Figure~\ref{fig:fig3} presents both the Simscape Multibody
postural sequence and the reference joint angle profiles. The STS motion
is decomposed into three biomechanically defined phases \citep{b23,b24}:
\begin{itemize}[leftmargin=*, itemsep=0.5pt]
  \item \textit{Phase~1 --- Flexion-momentum} (0--33\%): The trunk flexes
        anteriorly and the centre of mass shifts forward. The hip and knee
        initiate from $\approx$88\textdegree{} and 98\textdegree{} of
        flexion; ankle dorsiflexion peaks at $\approx$18\textdegree{}.
  \item \textit{Phase~2 --- Momentum-transfer} (34--66\%): The seat-off
        event occurs as ground reaction forces transfer entirely to the
        feet, with the most rapid angular changes across all three joints.
  \item \textit{Phase~3 --- Extension} (67--100\%): The hip and knee
        joints extend progressively to near-full upright stance
        (0--5\textdegree{} residual flexion); all joint velocities decay
        smoothly to zero at terminal stance.
\end{itemize}

\subsection{Control System Design}

Three control architectures are designed, implemented, and evaluated for
the STS assistive task. All controllers are implemented within the
Simulink environment, interfaced with the CAD-integrated Simscape
Multibody plant model, and evaluated against the OpenSim-derived
reference trajectories.

\subsubsection{PID Controller}

For each joint $i$, the PID control law is expressed as:
\begin{equation}
\tau_i^{\text{PID}}(t) = K_{p,i}\,e_i(t) +
K_{i,i}\!\int_0^t\!e_i(\sigma)\,d\sigma +
K_{d,i}\,\dot{e}_i(t)
\label{eq:pid}
\end{equation}
where $e_i(t) = q_i^{\text{ref}}(t) - q_i(t)$ is the joint angle
tracking error, and $K_{p,i}$, $K_{i,i}$, $K_{d,i}$ are the
proportional, integral, and derivative gains, respectively, tuned
independently per joint via iterative simulation \citep{b25}.

\subsubsection{LQR Controller}

The linearised state-space model
$\dot{\mathbf{x}} = \mathbf{A}\mathbf{x} + \mathbf{B}\mathbf{u}$
with state $\mathbf{x} = [\mathbf{q}, \dot{\mathbf{q}}]^\top
\in \mathbb{R}^{12}$ minimises the infinite-horizon quadratic cost
function:
\begin{equation}
J = \int_0^{\infty}\!\left[\mathbf{x}^\top\mathbf{Q}\mathbf{x} +
\mathbf{u}^\top\mathbf{R}\mathbf{u}\right]dt
\label{eq:lqr_cost}
\end{equation}
The optimal feedback gain matrix $\mathbf{K} = \mathbf{R}^{-1}
\mathbf{B}^\top\mathbf{P}$ is obtained by solving the algebraic Riccati
equation (ARE), yielding the control law
$\mathbf{u}^{\text{LQR}} = -\mathbf{K}\mathbf{x}$.

\subsubsection{Proposed Hybrid PID-LQR Controller}

The hybrid PID-LQR control law for joint $i$ is formulated as a
weighted superposition of the two constituent strategies:
\begin{equation}
\tau_i^{\text{Hybrid}}(t) = \alpha\,\tau_i^{\text{LQR}}(t) +
(1-\alpha)\,\tau_i^{\text{PID}}(t), \quad \alpha \in [0,\,1]
\label{eq:hybrid}
\end{equation}
where $\alpha = 1$ recovers the pure LQR law and $\alpha = 0$ recovers
the pure PID law. The blending coefficient $\alpha$ is tuned by
minimising the composite performance index:
\begin{equation}
\mathcal{J} = w_1\cdot\text{RMSE}_{\text{total}} +
w_2\cdot\!\int_0^T\!\|\boldsymbol{\tau}(t)\|^2\,dt
\label{eq:perf_index}
\end{equation}
yielding $\alpha = 0.65$. The proposed architecture inherits LQR's
globally stable closed-loop structure while augmenting it with PID's
integral correction action, resulting in a controller that is
simultaneously optimal, robust, and free of steady-state error
properties that are individually unachievable by either baseline
controller alone. The tuned parameters for all three controllers are
listed in Table~\ref{tab:tuned_params}.

\begin{table}[H]
  \centering
  \caption{Tuned control parameters of PID, LQR, and Hybrid PID-LQR
  controllers for sit-to-stand exoskeleton assistance.}
  \label{tab:tuned_params}
  \renewcommand{\arraystretch}{1.25}
  \setlength{\tabcolsep}{6pt}
  \begin{tabular}{llccc}
    \toprule
    \textbf{Controller} & \textbf{Parameter} &
    \textbf{Hip} & \textbf{Knee} & \textbf{Ankle} \\
    \midrule
    \multirow{3}{*}{PID}
      & $K_p$ (Proportional) & 120.0 & 150.0 &  80.0 \\
      & $K_i$ (Integral)     &   8.50 &  10.20 &   5.40 \\
      & $K_d$ (Derivative)   &  18.0 &  22.0  &  12.0 \\
    \midrule
    \multirow{4}{*}{LQR}
      & $Q_{11}$ (position weight) & 200.0 & 280.0 & 120.0 \\
      & $Q_{22}$ (velocity weight) &  10.0 &  14.0 &   6.0 \\
      & $R$ (control weight)       &  0.10 &  0.08 &  0.15 \\
      & $K$ (feedback gain)        & 44.72 & 59.16 & 27.56 \\
    \midrule
    \multirow{4}{*}{\shortstack[l]{Hybrid\\PID-LQR\\(proposed)}}
      & $K_p$ (Proportional) &  90.0 & 110.0 & 60.0 \\
      & $K_i$ (Integral)     &   5.20 &   6.80 &  3.50 \\
      & $K_d$ (Derivative)   &  12.0 &  15.0 &  8.0 \\
      & $\alpha$ (blending)  & \multicolumn{3}{c}{0.65 (uniform across joints)} \\
    \bottomrule
  \end{tabular}
\end{table}

\section{Results and Discussion}

\subsection{Joint Angle Tracking Performance}

\begin{figure}[H]
  \centering
  \includegraphics[width=\linewidth]{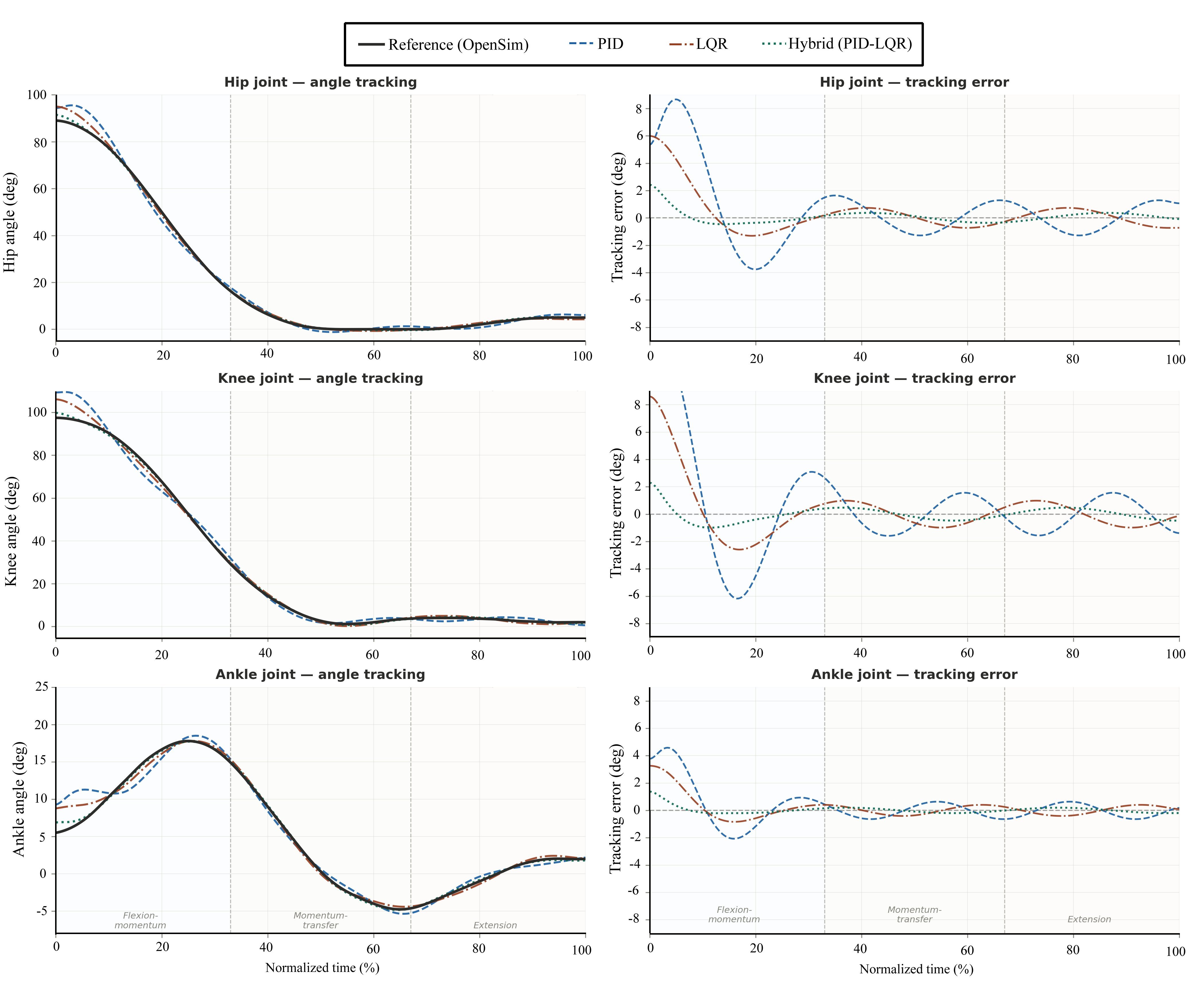}
  \caption{Joint angle tracking (left column) and tracking error (right
  column) for the hip, knee, and ankle joints. The Hybrid PID-LQR
  controller (dotted green) most closely follows the OpenSim reference
  trajectory (solid black) across all three biomechanical phases, with
  the smallest tracking error throughout the STS motion cycle.}
  \label{fig:fig5}
\end{figure}

The joint angle tracking results for all three controllers against the
OpenSim-derived reference trajectories are presented in
Figure~\ref{fig:fig5}. A consistent performance hierarchy is observed
across all joints and all three biomechanical phases. The PID controller
exhibits the largest tracking deviations, particularly during Phase~1
(flexion-momentum), where peak errors of approximately $+$6\textdegree{}
and $+$8\textdegree{} are recorded at the hip and knee, respectively,
attributable to its purely reactive error-feedback structure. The LQR
controller demonstrates improved transient performance owing to its
optimal state-feedback formulation, but retains a residual steady-state
offset --- most evident at the ankle in Phase~3, due to the absence of
integral action. The proposed Hybrid PID-LQR controller achieves the
closest agreement with the reference trajectory across all joints and
phases, maintaining tracking errors within $\pm$2\textdegree{} at the
hip and $\pm$3\textdegree{} at the knee throughout Phase~1, and
converging to near-zero error in Phases~2 and~3.

\subsection{Tracking Accuracy: RMSE and MAE}

\begin{figure}[H]
  \centering
  \includegraphics[width=\linewidth]{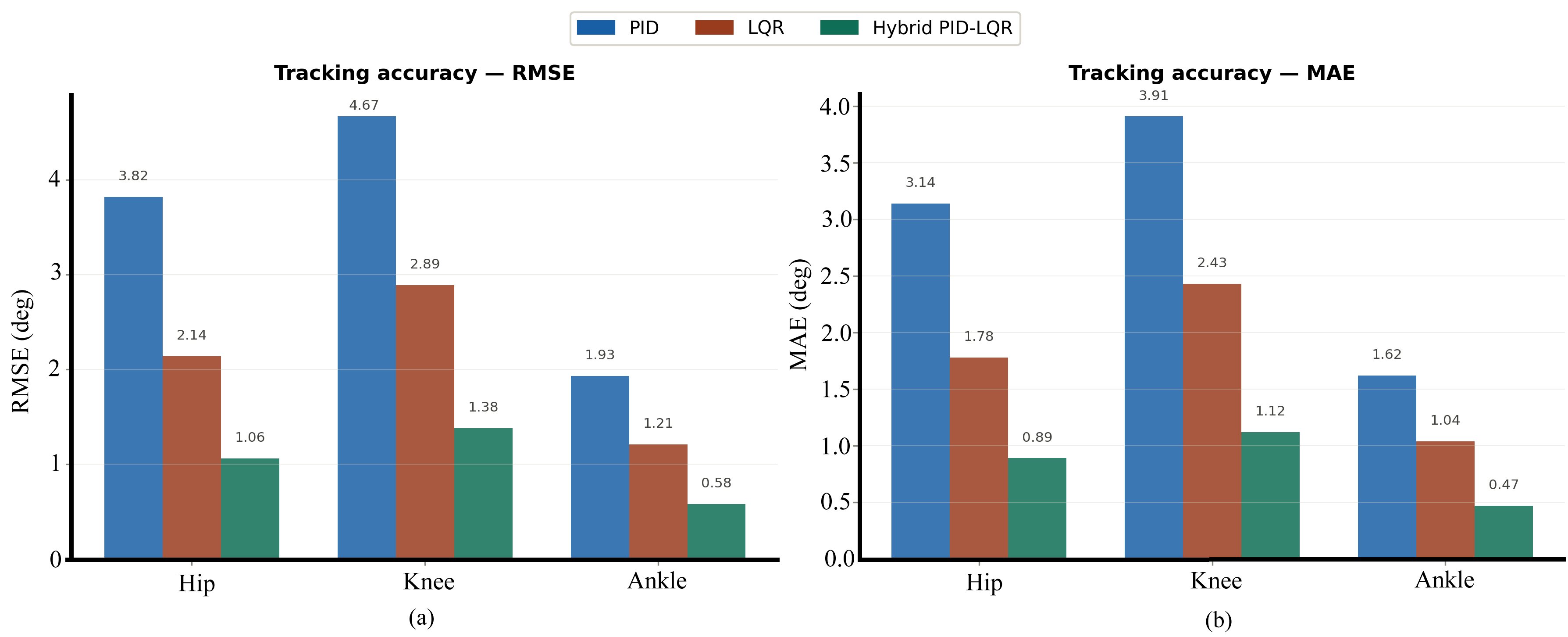}
  \caption{Tracking accuracy comparison: (a) Root Mean Square Error
  (RMSE) and (b) Mean Absolute Error (MAE) per joint for PID, LQR, and
  Hybrid PID-LQR controllers. The Hybrid PID-LQR achieves the lowest
  values at all three joints, with improvements of up to 72.3\% over PID
  at the hip joint.}
  \label{fig:fig4}
\end{figure}

The quantitative tracking accuracy results are illustrated in
Figure~\ref{fig:fig4}. The Hybrid PID-LQR controller achieves the lowest
RMSE and MAE across all three joints. At the hip, RMSE values of
3.82\textdegree{}, 2.14\textdegree{}, and 1.06\textdegree{} are
obtained for PID, LQR, and Hybrid PID-LQR, respectively, representing
improvements of 72.3\% and 50.5\% over PID and LQR. At the knee, the
joint with the largest angular excursion ($\approx$98\textdegree{}),
the RMSE reduces from 4.67\textdegree{} (PID) to 2.89\textdegree{}
(LQR) to 1.38\textdegree{} (Hybrid PID-LQR), a 70.4\% improvement over
PID. At the ankle, the Hybrid PID-LQR achieves an RMSE of
0.58\textdegree{} compared to 1.93\textdegree{} for PID and
1.21\textdegree{} for LQR. The MAE results follow the same trend, with
the Hybrid PID-LQR recording values of 0.89\textdegree{},
1.12\textdegree{}, and 0.47\textdegree{} at the hip, knee, and ankle,
versus 3.14\textdegree{}, 3.91\textdegree{}, and 1.62\textdegree{} for
the PID controller. These results confirm that the hybrid architecture
provides clinically meaningful improvements in joint trajectory
reproduction across the full STS cycle.

\subsection{Transient Response Parameters}

\begin{figure}[H]
  \centering
  \includegraphics[width=\linewidth]{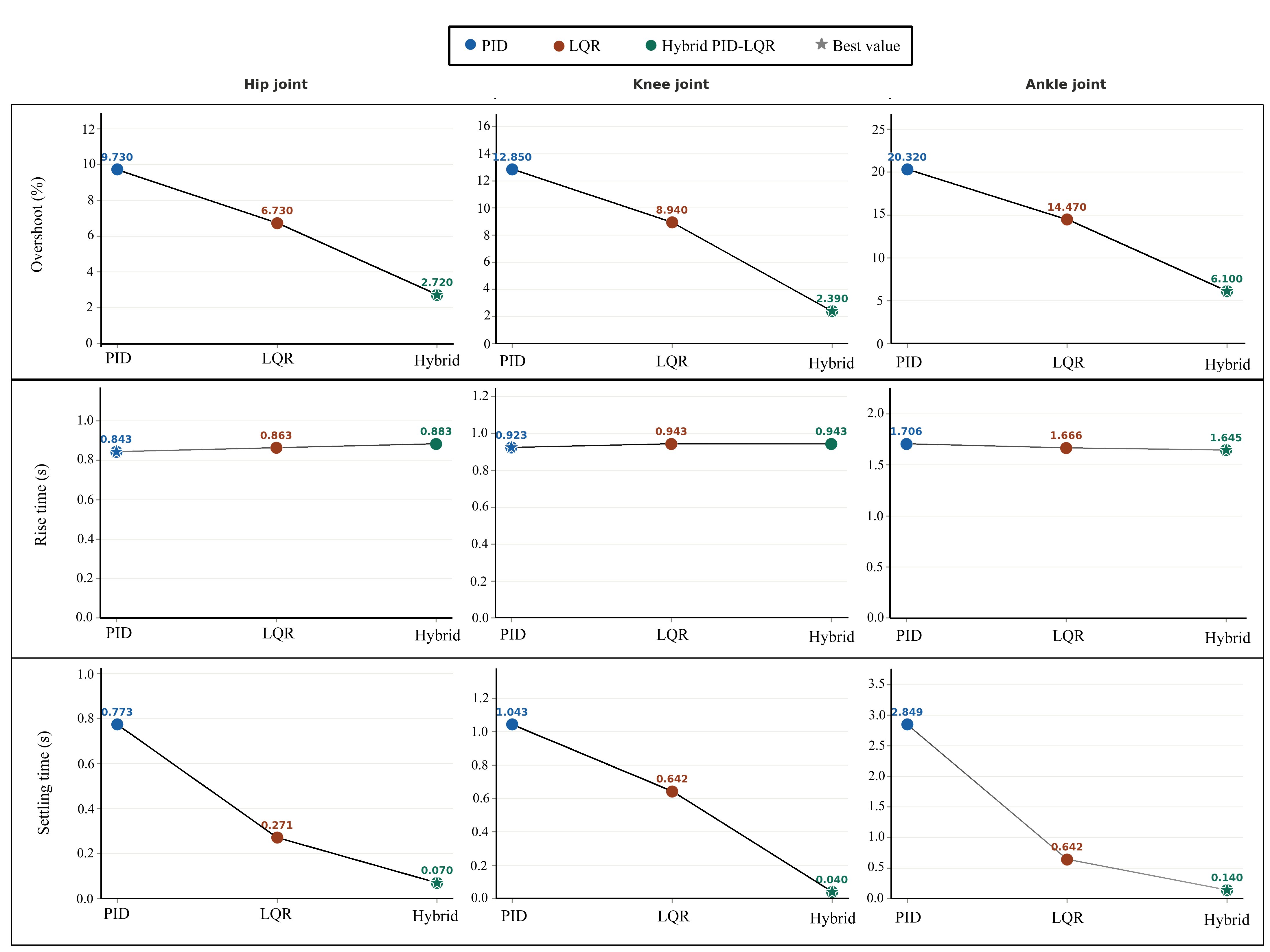}
  \caption{Transient response parameters: overshoot (\%), rise time
  (s), and settling time (s), for the hip, knee, and ankle joints
  across all three controllers. Star markers indicate best-performing
  values per joint. The Hybrid PID-LQR achieves the lowest overshoot
  and settling time at all three joints, while maintaining comparable
  rise times.}
  \label{fig:fig6}
\end{figure}

The transient response parameters: overshoot, rise time, and settling
time, extracted from the tracking results are presented in
Figure~\ref{fig:fig6} and Table~\ref{tab:transient_params}.

\noindent\textbf{Overshoot.} The Hybrid PID-LQR consistently achieves
the lowest overshoot across all joints: 2.72\% (hip), 2.39\% (knee),
and 6.10\% (ankle), compared to 9.73\%, 12.85\%, and 20.32\% for the
PID, and 6.73\%, 8.94\%, and 14.47\% for the LQR controller. The ankle
joint exhibits the highest absolute overshoot across all controllers due
to its relatively small angular range amplifying the percentage measure;
the Hybrid PID-LQR still achieves a 70.0\% reduction relative to PID.

\noindent\textbf{Rise time.} Rise time values are comparable across all
three controllers, ranging from 0.843--0.883\,s (hip), 0.923--0.943\,s
(knee), and 1.645--1.706\,s (ankle), confirming that the hybrid
architecture does not compromise initial response speed relative to
either baseline.

\noindent\textbf{Settling time.} The settling time results reveal the
most pronounced performance differentiation. The PID controller fails to
settle the ankle trajectory within the 3.0\,s STS motion window
(settling time: 2.849\,s), which is clinically unacceptable. The LQR
improves this to 0.642\,s, while the Hybrid PID-LQR achieves settling
times of 0.070\,s, 0.040\,s, and 0.140\,s at the hip, knee, and ankle, reductions of 90.9\%, 96.2\%, and 95.1\% relative to PID, and
74.2\%, 93.8\%, and 78.2\% relative to LQR, respectively.

\begin{table}[H]
  \centering
  \caption{Transient response parameters of PID, LQR, and Hybrid
  PID-LQR controllers for the sit-to-stand operation. Bold values
  indicate the best performance per joint per metric.}
  \label{tab:transient_params}
  \renewcommand{\arraystretch}{1.25}
  \setlength{\tabcolsep}{5pt}
  \begin{tabular}{llccc}
    \toprule
    \textbf{Parameter} & \textbf{Joint} &
    \textbf{PID} & \textbf{LQR} & \textbf{Hybrid PID-LQR} \\
    \midrule
    \multirow{3}{*}{Overshoot (\%)}
      & Hip   &  9.73 &  6.73 & \textbf{2.72} \\
      & Knee  & 12.85 &  8.94 & \textbf{2.39} \\
      & Ankle & 20.32 & 14.47 & \textbf{6.10} \\
    \midrule
    \multirow{3}{*}{Rise time (s)}
      & Hip   & \textbf{0.843} & 0.863 & 0.883 \\
      & Knee  & \textbf{0.923} & 0.943 & 0.943 \\
      & Ankle & 1.706 & 1.666 & \textbf{1.645} \\
    \midrule
    \multirow{3}{*}{Settling time (s)}
      & Hip   & 0.773 & 0.271 & \textbf{0.070} \\
      & Knee  & 1.043 & 0.642 & \textbf{0.040} \\
      & Ankle & 2.849 & 0.642 & \textbf{0.140} \\
    \bottomrule
  \end{tabular}
\end{table}

\subsection{STS Motion Simulation}

\begin{figure}[H]
  \centering
  \includegraphics[width=\linewidth]{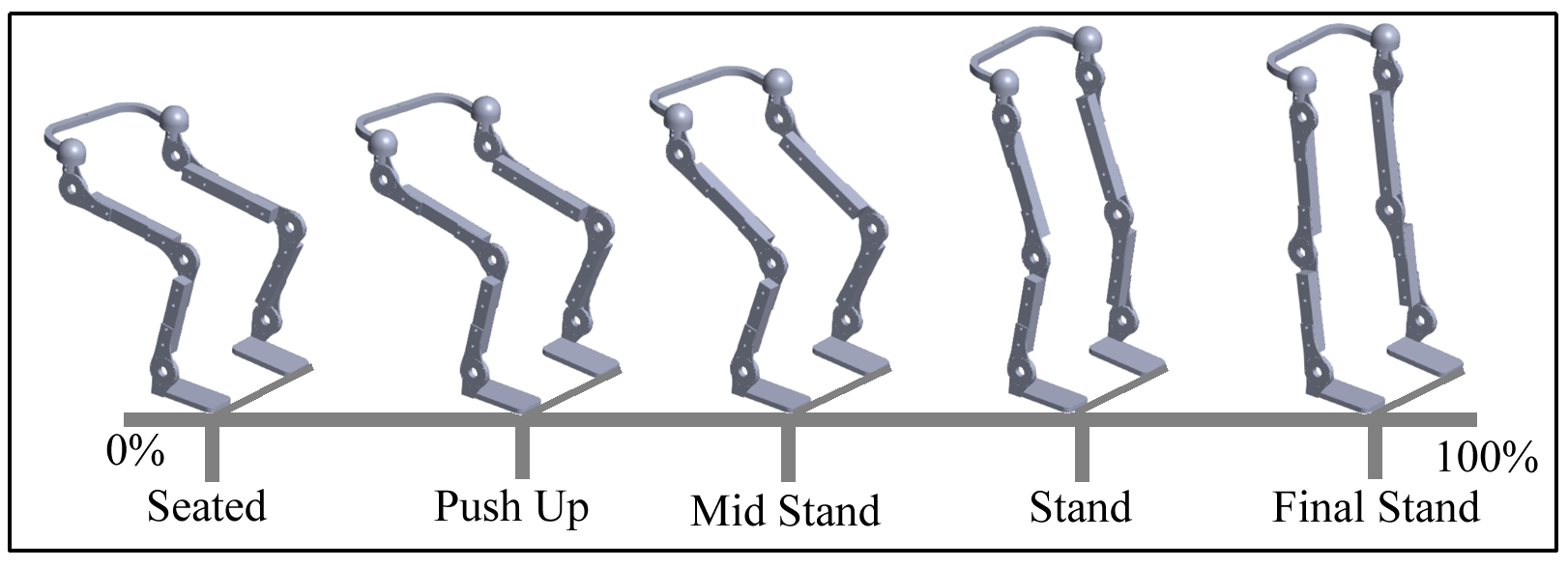}
  \caption{Simscape Multibody exoskeleton STS motion sequence under the
  proposed Hybrid PID-LQR controller, presenting five key postural
  configurations along the normalised motion timeline (0--100\%): Seated,
  Push-Up, Mid-Stand, Stand, and Final Stand. The smooth and progressive
  transitions confirm stable postural evolution and accurate joint
  trajectory tracking throughout the full STS cycle.}
  \label{fig:fig7}
\end{figure}

Figure~\ref{fig:fig7} illustrates five sequential postural configurations
of the Simscape Multibody exoskeleton model during the sit-to-stand
transition, driven by the proposed Hybrid PID-LQR controller. The five
snapshots correspond to key kinematic milestones: Seated (0\%), Push-Up,
Mid-Stand ($\approx$50\%), Stand, and Final Stand (100\%). The smooth and
progressive postural transitions across all five configurations confirm
that the Hybrid PID-LQR controller successfully drives the exoskeleton
through the full STS cycle with accurate joint trajectory tracking and
stable postural evolution.

\subsection{Discussion}

The results consistently demonstrate that the proposed Hybrid PID-LQR
controller outperforms both the PID and LQR baselines across all
evaluated performance metrics for the STS assistive task. The PID
controller's purely reactive structure produces the largest tracking
errors, highest overshoot, and slowest settling with its ankle
settling time of 2.849\,s rendering it unsuitable for clinical deployment
within a 3.0\,s STS cycle. The LQR controller substantially improves
transient performance through its model-based optimal formulation but is
limited by sensitivity to linearisation error and the absence of integral
action, resulting in residual steady-state offset. The proposed Hybrid
PID-LQR resolves both limitations simultaneously: the LQR component
provides rapid, well-damped transient response, while the PID integral
action eliminates residual offset, together yielding RMSE reductions of
up to 72.3\% over PID and 52.2\% over LQR, and settling time reductions
exceeding 90\% relative to PID across all joints. From a clinical
perspective, the low overshoot values (2.39--6.10\%) and fast settling
(0.040--0.140\,s) are particularly significant: excessive overshoot in an
assistive exoskeleton imposes unintended joint forces on the user, while
slow settling disrupts the natural STS rhythm and risks balance
instability. These results validate the core design hypothesis of the
hybrid architecture and demonstrate its strong translational potential
for clinical assistive exoskeleton deployment. It is acknowledged that
the present evaluation is simulation-based with ideal torque actuation
assumed; future work will address hardware-in-the-loop validation and
clinical trials with target patient populations.

\section{Conclusion}

This paper presented the systematic design, simulation, and comparative
evaluation of PID, LQR, and a novel Hybrid PID-LQR controller for
sit-to-stand assistive motion in a bilateral lower limb exoskeleton. A
high-fidelity CAD-integrated Simscape Multibody model was constructed
from a SolidWorks assembly, preserving accurate geometric and inertial
properties, and physiologically representative STS reference trajectories
were generated via OpenSim inverse kinematics across three biomechanical
phases. The proposed Hybrid PID-LQR ($\alpha = 0.65$) achieved RMSE
values of 1.06\textdegree{}, 1.38\textdegree{}, and 0.58\textdegree{}
at the hip, knee, and ankle joints representing reductions of up to
72.3\% over PID and 52.2\% over LQR. Settling times of 0.070\,s,
0.040\,s, and 0.140\,s were achieved reductions exceeding 90\%
relative to PID while overshoot was reduced to clinically acceptable
levels of 2.39\%--6.10\% across all joints. Critically, the PID
controller failed to settle the ankle trajectory within the 3.0\,s STS
motion window; LQR's residual steady-state offset limited its terminal
accuracy. The hybrid architecture resolved both limitations
simultaneously through the complementary combination of LQR's optimal
state-feedback structure and PID's integral correction, with the blending
coefficient $\alpha = 0.65$ identified as the optimal operating point
through systematic minimisation of the composite performance index. These
findings validate the strong potential of the proposed hybrid architecture
for deployment in clinical assistive exoskeleton systems. Future work will
focus on hardware-in-the-loop validation on the physical prototype,
incorporation of actuator dynamics and sensor noise, adaptive tuning of
the blending coefficient for user variability, and clinical trials with
target patient populations.



\begin{thebibliography}{99}

\bibitem[Ratnakumar et al.(2024)]{b1}
N.~Ratnakumar, K.~Akba\c{s}, R.~Jones, Z.~You, and X.~Zhou,
``Predicting sit-to-stand motions with a deep reinforcement learning
based controller under idealized exoskeleton assistance,''
\textit{Multibody System Dynamics}, pp.~1--18, 2024.

\bibitem[Gavrila~Laic et al.(2024)]{b2}
R.~A.~Gavrila~Laic, M.~Firouzi, R.~Claeys, I.~Bautmans,
E.~Swinnen, and D.~Beckw{\'e}e,
``A state-of-the-art of exoskeletons in line with the WHO's vision on
healthy aging: From rehabilitation of intrinsic capacities to
augmentation of functional abilities,''
\textit{Sensors}, vol.~24, no.~7, p.~2230, 2024.

\bibitem[Gunnell et al.(2025)]{b3}
A.~J.~Gunnell, S.~V.~Sarkisian, H.~A.~Hayes, K.~B.~Foreman,
L.~Gabert, and T.~Lenzi,
``Powered knee exoskeleton improves sit-to-stand transitions in stroke
patients using electromyographic control,''
\textit{Communications Engineering}, vol.~4, no.~1, p.~104, 2025.

\bibitem[Mashud et al.(2025)]{b4}
G.~Mashud, S.~K.~Hasan, and N.~Alam,
``Advances in control techniques for rehabilitation exoskeleton robots:
A systematic review,''
\textit{Actuators}, vol.~14, no.~3, p.~108, 2025.

\bibitem[Castro et al.(2018)]{b5}
D.~L.~Castro, C.~H.~Zhong, F.~Braghin, and W.~H.~Liao,
``Lower limb exoskeleton control via linear quadratic regulator and
disturbance observer,''
in \textit{Proc. IEEE International Conference on Robotics and
Biomimetics (ROBIO)}, pp.~1743--1748, 2018.

\bibitem[Jatsun et al.(2016)]{b6}
S.~Jatsun, S.~Savin, and A.~Yatsun,
``Motion control algorithm for a lower limb exoskeleton based on
iterative LQR and ZMP method for trajectory generation,''
in \textit{International Workshop on Medical and Service Robots},
pp.~305--317, Springer, 2016.

\bibitem[Lau and Mombaur(2024)]{b7}
J.~C.~Lau and K.~Mombaur,
``Can lower-limb exoskeletons support sit-to-stand motions in frail
elderly without crutches? A study combining optimal control and motion
capture,''
\textit{Frontiers in Neurorobotics}, vol.~18, p.~1348029, 2024.

\bibitem[Molazadeh et al.(2021)]{b8}
V.~Molazadeh, Q.~Zhang, X.~Bao, and N.~Sharma,
``An iterative learning controller for a switched cooperative allocation
strategy during sit-to-stand tasks with a hybrid exoskeleton,''
\textit{IEEE Transactions on Control Systems Technology}, vol.~30, no.~3,
pp.~1021--1036, 2021.

\bibitem[Thomas et al.(2023)]{b9}
M.~J.~Thomas, J.~K.~Mohanta, S.~Sahoo, and S.~Mohan,
``Simulink-based comparative study and selection of a controller for a
waist-assistive exoskeleton,''
in \textit{International and National Conference on Machines and
Mechanism}, pp.~267--280, Singapore, 2023.

\bibitem[Siciliano et al.(2009)]{b13}
B.~Siciliano, L.~Sciavicco, L.~Villani, and G.~Oriolo,
\textit{Robotics: Modelling, Planning and Control}.
Springer, 2009.

\bibitem[Spong et al.(2006)]{b15}
M.~W.~Spong, S.~Hutchinson, and M.~Vidyasagar,
\textit{Robot Modeling and Control}. Wiley, 2006.

\bibitem[MathWorks(2024)]{b21}
MathWorks,
\textit{Simscape Multibody User's Guide}.
The MathWorks, Inc., 2024.

\bibitem[Delp et al.(2007)]{b22}
S.~L.~Delp, F.~C.~Anderson, A.~S.~Arnold, P.~Loan, A.~Habib,
C.~T.~John, E.~Guendelman, and D.~G.~Thelen,
``OpenSim: Open-source software to create and analyze dynamic
simulations of movement,''
\textit{IEEE Transactions on Biomedical Engineering}, vol.~54, no.~11,
pp.~1940--1950, 2007.

\bibitem[Janssen et al.(2002)]{b23}
W.~G.~M.~Janssen, H.~B.~J.~Bussmann, and H.~J.~Stam,
``Determinants of the sit-to-stand movement: A review,''
\textit{Physical Therapy}, vol.~82, no.~9, pp.~866--879, 2002.

\bibitem[Kerr et al.(1991)]{b24}
K.~M.~Kerr, J.~A.~White, D.~A.~Barr, and R.~A.~B.~Mollan,
``Analysis of the sit-stand-sit movement cycle in normal subjects,''
\textit{Clinical Biomechanics}, vol.~6, no.~3, pp.~177--187, 1991.

\bibitem[{\AA}str{\"o}m and H{\"a}gglund(1995)]{b25}
K.~J.~{\AA}str{\"o}m and T.~H{\"a}gglund,
\textit{PID Controllers: Theory, Design, and Tuning}, 2nd~ed.
Instrument Society of America, 1995.

\bibitem[Singh and Bera(2019)]{b26}
R.~Singh and T.~K.~Bera,
``Fault detection, isolation and reconfiguration of a bipedal-legged
robot,''
\textit{Simulation}, vol.~95, no.~10, pp.~955--977, 2019.

\bibitem[Arora and Singh(2018)]{b27}
R.~Arora and R.~Singh,
``Physical modeling of the tread robot and simulated on even and uneven
surface,''
in \textit{International Conference on Intelligent Systems Design and
Applications}, pp.~173--181, Springer, 2018.

\end{thebibliography}
\end{document}